\documentclass[sn-basic,iicol]{sn-jnl}

\usepackage{graphicx}%
\usepackage{multirow}%
\usepackage{amsmath,amssymb,amsfonts}%
\usepackage{amsthm}%
\usepackage{mathrsfs}%
\usepackage[title]{appendix}%
\usepackage{xcolor}%
\usepackage{textcomp}%
\usepackage{manyfoot}%
\usepackage{booktabs}%
\usepackage{algorithm}%
\usepackage{algorithmicx}%
\usepackage{algpseudocode}%
\usepackage{listings}%
\usepackage{makecell}
\usepackage{subfigure}
\usepackage{multicol}
\usepackage{booktabs}

\usepackage{todonotes}
\usepackage{xcolor}

\theoremstyle{thmstyleone}%
\theoremstyle{thmstyletwo}%

\theoremstyle{thmstylethree}%

\begin{document}

\title[Exploring Vision-Language Models for Imbalanced Learning]{Exploring Vision-Language Models for Imbalanced Learning}



\author{\fnm{Yidong} \sur{Wang}$^\text{1}$}

\author{\fnm{Zhuohao} \sur{Yu}$^\text{1}$}

\author{\fnm{Jindong} \sur{Wang}$^\text{2}$}\email{jindong.wang@microsoft.com}

\author{\fnm{Qiang} \sur{Heng}$^\text{3}$}

\author{\fnm{Hao} \sur{Chen}$^\text{4}$}

\author{\fnm{Wei} \sur{Ye}$^\text{1}$}\email{wye@pku.edu.cn}

\author{\fnm{Rui} \sur{Xie}$^\text{1}$} 

\author{\fnm{Xing} \sur{Xie}$^\text{2}$}

\author{\fnm{Shikun} \sur{Zhang}$^\text{1}$}\email{zhangsk@pku.edu.cn}

\affil[1]{\orgdiv{National Engineering Research Center for Software Engineering}, \orgname{Peking University}}

\affil[2]{\orgname{Mircosoft Research Asia}}

\affil[3]{\orgname{North Carolina State University}}

\affil[4]{\orgname{Carnegie Mellon University}}















\abstract{Vision-Language models (VLMs) that use contrastive language-image pre-training have shown promising zero-shot classification performance. However, their performance on imbalanced dataset is relatively poor, where the distribution of classes in the training dataset is skewed, leading to poor performance in predicting minority classes. For instance, CLIP achieved only 5\% accuracy on the iNaturalist18 dataset. We propose to add a lightweight decoder to VLMs to avoid OOM (out of memory) problem caused by large number of classes and capture nuanced features for tail classes. Then, we explore improvements of VLMs using prompt tuning, fine-tuning, and incorporating imbalanced algorithms such as Focal Loss, Balanced SoftMax and Distribution Alignment. Experiments demonstrate that the performance of VLMs can be further boosted when used with decoder and imbalanced methods. Specifically, our improved VLMs significantly outperforms zero-shot classification by an average accuracy of \textbf{6.58}\%, \textbf{69.82}\%, and \textbf{6.17}\%, on ImageNet-LT, iNaturalist18, and Places-LT, respectively. We further analyze the influence of pre-training data size, backbones, and training cost. Our study highlights the significance of imbalanced learning algorithms in face of VLMs pre-trained by huge data. We release our code at \url{https://github.com/Imbalance-VLM/Imbalance-VLM}.}




\maketitle

\section{Introduction}
\label{sec1}

Vision-Language Models (VLMs) using joint language-image pre-training have quickly gained popularity due to their impressive performance in a wide range of computer vision tasks~\citep{radford2021learning,schuhmannlaion,yu2022coca,luddecke2022image}. One notable application of these models is the zero-shot classification where the model recognizes objects or scenes from a set of classes it has never seen before.
VLMs achieve this by associating textual descriptions with images during prediction, allowing them to perform well on unseen data without the need for additional training on the target classes. 

\begin{figure*}[htbp]
	\centering
	\subfigure[Class distribution]{
		\includegraphics[width=0.6\textwidth]{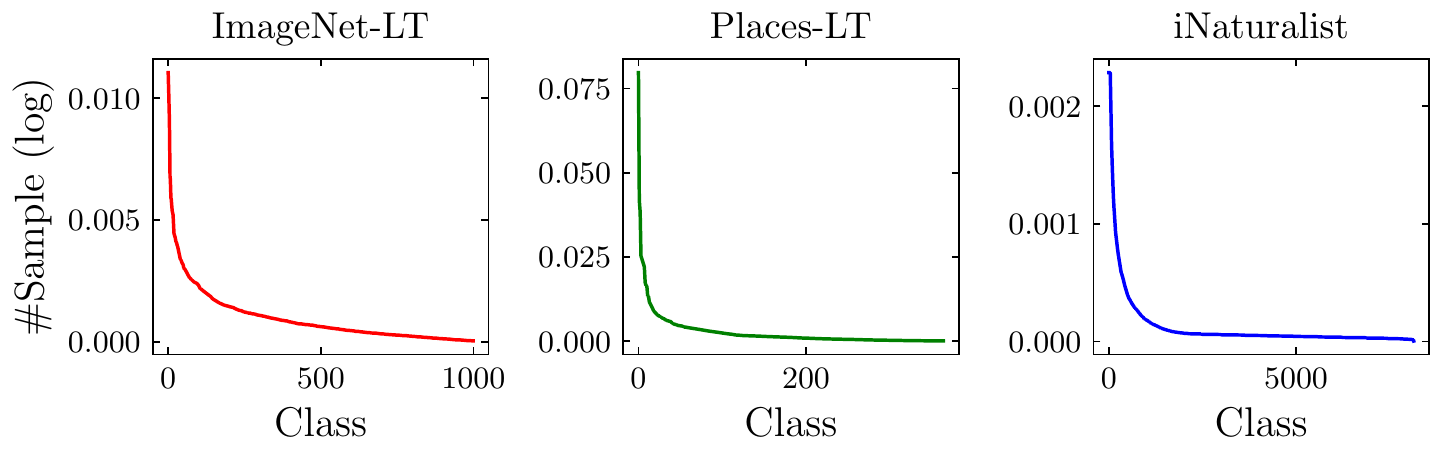}
  \label{fig-dist}
	}
	\subfigure[Overall accuracy]{
		\label{fig:demo1_2}
		\includegraphics[width=0.35\textwidth]{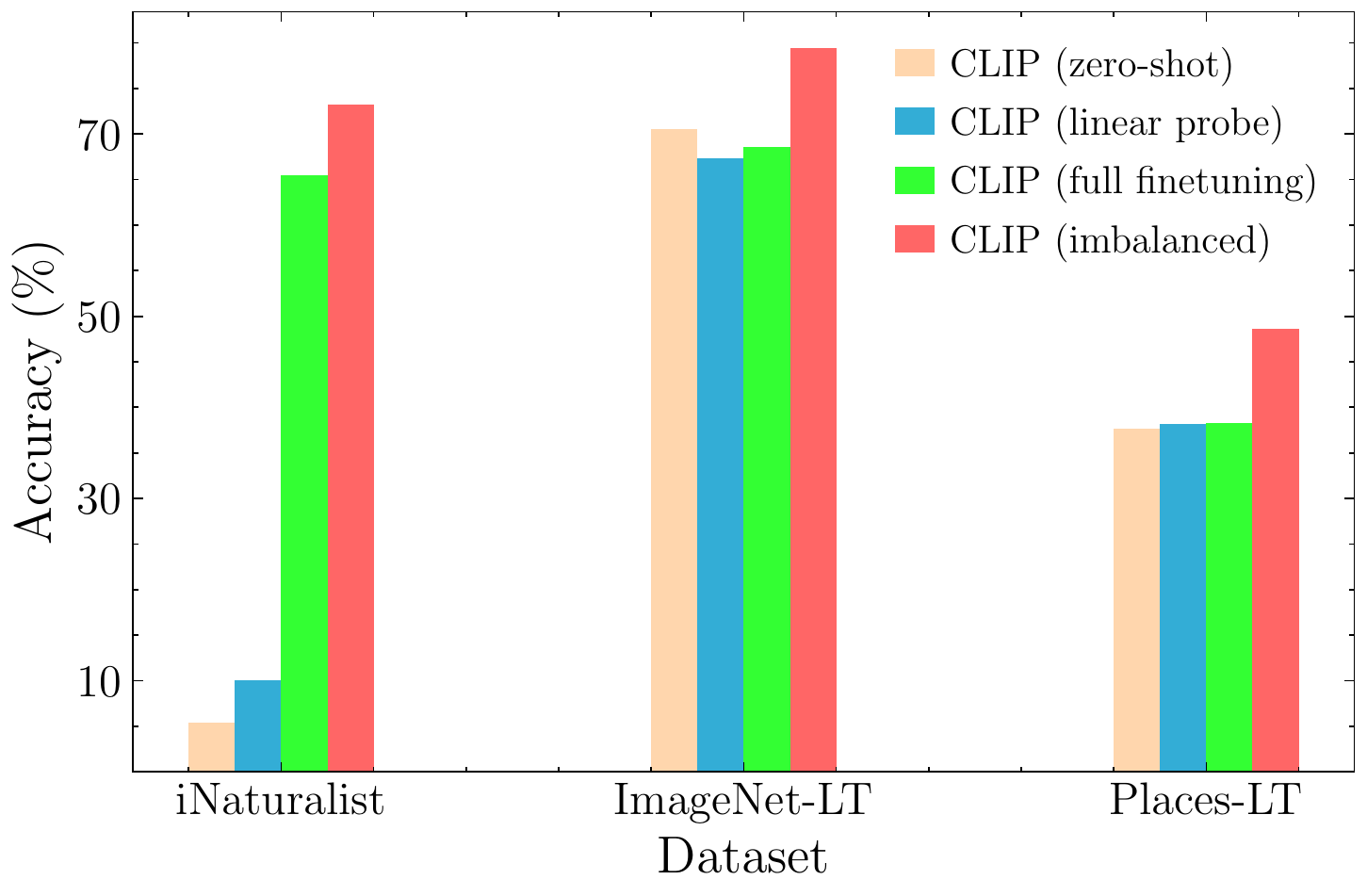}
  \label{fig-comp}
	}
	\caption{(a) Class distribution of ImageNet-LT, Places-LT and iNaturalist18. Number of samples of each class is transformed using logarithm. (b) Overall accuracy of zero-shot, fine-tuned, and imbalanced methods with CLIP~\citep{radford2021learning}.}
	\label{fig-intro}
\end{figure*}

However, despite their promising performance on zero-shot classification, we have empirically observed that these models cannot perform perfectly on \emph{imbalanced} datasets where high-frequency classes, i.e., the head or common classes, contain most of the instances, and low-frequency classes, i.e., tail or rare classes, have very few instances.
Note that the test set is often balanced to ensure fairness for each class during evaluation.
The zero-shot classification performance of VLMs on imbalanced datasets is limited due to several factors, including the inherent bias in the pre-training data, the lack of exposure to the tail classes during pre-training, and the lack of techniques to mitigate the effects of class imbalance~\citep{schuhmannlaion}.
As a result, VLMs often perform poorly on tail classes, which can be critical for many applications related to safety or health, such as autonomous driving and medical diagnosis~\citep{yang2020rethinking}.
Hence, it is intuitive to ask: \emph{is this long-tailed pre-training style actually transfer or influence downstream long-tailed classes?}

As shown in \figurename~\ref{fig-comp}, on the iNaturalist18 dataset~\citep{van2018inaturalist}, the zero-shot performance of CLIP~\citep{radford2021learning}, one of the most popular VLMs, is notably poor.
Is it possible to improve their performance for imbalanced tasks?
We show that when VLMs are combined with supervised fine-tuning using imbalanced methods, they show a considerable improvement in accuracy.
On other datasets such as ImageNet-LT and Places-LT, zero-shot VLMs perform relatively well, but the accuracy can further be boosted when combined with supervised training, showing the potential of fine-tuning VLMs in imbalanced settings.




In this paper, we explore the use of supervised imbalanced methods in conjunction with VLMs to improve the performance of VLMs on tail classes, which could fully unleash the power of VLMs.\footnote{We mainly deal with a supervised imbalanced setting, i.e., the training data is fully labeled. There are other new emerging areas such as semi-supervised imbalanced learning~\citep{chen2022embarrassingly} and they are out of the scope of this paper.}
Specifically, we propose to incorporate a lightweight \emph{decoder} after the Vision Transformer (ViT) of VLMs to save memory and capture subtle features for tail classes.
Based on the modification, we investigate several class-balanced loss function engineering methods and two-stage methods to improve the performance of VLMs on imbalanced datasets.
Some class-balanced loss functions adjust the logits instead of weighting the losses~\citep{menon2020long,cao2019learning,ren2020balanced} to achieve more balanced gradients between classes.
Recently, two-stage methods that re-adjust the classifier with the fixed representation backbone trained using standard Cross-Entropy Loss with instance-balanced sampling show strong performance compared with previous loss function methods~\citep{kang2019decoupling,zhang2021distribution,wang2022margin}.
Our experimental results demonstrate that the performance of VLMs can be further improved when used with decoder and imbalanced methods.
Our findings suggest that the combination of VLMs and imbalanced methods can be effective in addressing the challenge of class imbalance.


The contributions of this paper are as follows:

\begin{enumerate}
    \item To our best knowledge, we are the \emph{first} to provide a thorough exploration of combining vision-language models with imbalanced classification methods, providing rich experience for research on this topic.
    \item Our insightful analysis shows that VLMs could show poor performance on some imbalanced datasets, indicating that there is still large room for improvement.
    \item We propose simple modifications of VLMs to improve their performance using existing imbalanced learning methods, showing the possibility to further enhance VLMs for imbalanced learning. Experiments show that our modification with imbalanced algorithms significantly outperforms the zero-shot performance by \textbf{6.58}\%, \textbf{69.82}\%, and \textbf{6.17}\% on ImageNet-LT, iNaturalist18, and Places-LT datasets, respectively.
    \item Our training code, details, and benchmarks are open-sourced at \url{https://github.com/Imbalance-LVM/Imbalance-LVM}, which can facilitate future research on this topic on using VLMs for imbalanced learning.

\end{enumerate}

\section{Related work}
\subsection{Vision Foundation Models}
The trend towards increasingly larger vision models has been observed in recent years, resulting in state-of-the-art performance on a variety of computer vision tasks~\citep{dehghani2023scaling,liu2022swin}.
In this work, we focus on Large Joint Vision-Language Pre-training Vision Models which enables natural language supervision for vision models.
Instead of training in a fixed set of predetermined object categories, CLIP~\citep{radford2021learning} directly learning from raw text about images and broaden the source of supervision. By adopting natural language to reference learned visual concepts or describe new ones, clip enables zero-shot classification on unseen tasks and is competitive with a fully supervised baseline on many tasks, without any dataset-specific training. BLIP~\citep{li2022blip} then adopt caption bootstrapping to leverage noisy web data in an efficient and effective manner. By generating synthetic captions using a captioner and filtering the noisy ones from a large noisy image-text pairs dataset collected from the web, BLIP can achieved great improvement on zero shot classification tasks. The success of CLIP and BLIP is inseparable from large amounts of high-quality datasets. To facilitate the research for LVMs, LAION-5B~\citep{schuhmannlaion} is made public, which is a dataset consisting of 5.85 billion CLIP-filtered image-text pairs, of which 2.32 billion contain English language. Recently, several work also investigate imbalanced recognition using CLIP~\citep{ma2021simple} and introduce text modality for long-tailed recognition tasks~\citep{tian2022vl}. Other work seek to enhance the performance of Vision Transformers (ViT) on imbalanced datasets by integrating unsupervised learning~\citep{xu2023rethink}. However, their training objectives typically follow the contrastive learning objective of CLIP or the mask image modeling of MAE~\citep{he2022masked}, while our focus is combining VLMs with imbalanced learning methods. 

\subsection{Imbalanced Learning}

Imbalanced classification has attracted significant interest recently as it is a common issue in the real world~\citep{yang2022survey,tang2020long,equal,wang2021seesaw,wei2022open}. Recent approaches can be classified into four categories.


\textbf{Loss function engineering.} Loss function engineering aims to obtain balanced gradients during training. This technique includes loss re-weighting and logits adjustment. Loss re-weighting adjusts the weights of losses for different classes or instances to achieve a more balanced distribution~\citep{byrd2019effect,khan2017cost,wang2017learning}. Instances from tail classes are assigned larger weights than those from head classes. On the other hand, logits adjustment methods adjust the logits to obtain balanced gradients during training without re-weighting losses~\citep{menon2020long,cao2019learning,ren2020balanced,yang2009margin}.

\textbf{Two-stage Decision boundary adjustment.} The data re-sampling and loss function engineering methods can have a side impact on data representations when training data is imbalanced~\citep{ren2020balanced,zhang2021distribution,wang2022margin}. In order to mitigate this impact, decision boundary adjustment techniques are employed to re-adjust the classifier head in a learnable manner~\citep{platt1999large,kang2019decoupling,zhang2021distribution,wang2022margin} after standard training. 

\textbf{Other methods.} Other paradigms include task-specific architecture design~\citep{wang2021contrastive,zhou2020bbn,wang2021rsg}, transfer learning~\citep{liu2019large,yin2019feature}, domain adaptation~\citep{jamal2020rethinking}, semi-supervised learning, and self-supervised learning~\citep{yang2020rethinking} which demand non-trivial architecture design or external data.

\section{Are Vision-Language Models All We Need for Imbalanced Learning?}
\label{sec-zeroshot}

The huge success of vision-language models on different standard tasks~\citep{radford2021learning, yu2022coca} naturally motivates a question: \emph{Are vision-language models all we need for imbalanced learning?}
We try to answer this question by exploring the supervised fine-tuning of VLMs.

\subsection{Imbalanced Learning}

We use imbalanced classification as the main task, where the training data distribution is imbalanced and the test data distribution is balanced.
Formally, Let $\mathcal{D}=\{(\mathbf{x}_i,y_i)\}_{i=1}^n$ be a collection of training data with length $n$, where $y_i$ denotes the label of sample $\mathbf{x}_i$.
The number of samples in each class $j$ can be denoted by $n_j$ and the number of classes is $K$.
Without loss of generality, we assume $n_1 > n_2 > \cdots >n_K$ in the imbalanced setting.
In supervised learning, the prediction function of a neural model consists of a feature learning function $f: \mathbf{x} \mapsto \mathbf{z}$ and a linear classifier $g: \mathbf{z} \mapsto \mathbf{y}$, where $\mathbf{z} \in \mathbb{R}^{p}$ is the $p$-dimensional representation.
The logit of class $j$ is computed as:
\begin{equation}
\eta_j = g(\mathbf{z}) := \mathbf{W}_j \mathbf{z} + \mathbf{b}_j,
\end{equation}
where $\mathbf{W}_j$ and $\mathbf{b}_j$ are the weight matrix and bias in the linear classifier, respectively.
Then, the classification probability of $\mathbf{x}_i$ is computed as:
\begin{equation}
\label{eq:softmax}
    p(y=y_i|\mathbf{x}_i;\theta_r,\theta_c) = \frac{\exp(\eta_{y_i})}{\sum_{j=1}^K  \exp(\eta_j)},
\end{equation}
where $\theta_r$ and $\theta_c$ represent the trainable parameters of feature representation learning function $f$ and the linear classifier $g$. The training loss is typically computed as the cross-entropy loss:
\begin{equation}
    \label{eq:loss_softmax}
    \ell(\mathbf{x}_i, y_i;\theta_r,\theta_c) = -\log p(y=y_i|\mathbf{x}_i;\theta_r,\theta_c).
\end{equation}

\begin{figure*}[t!]%
\centering
\vspace{-.2in}
\includegraphics[width=0.9\textwidth]{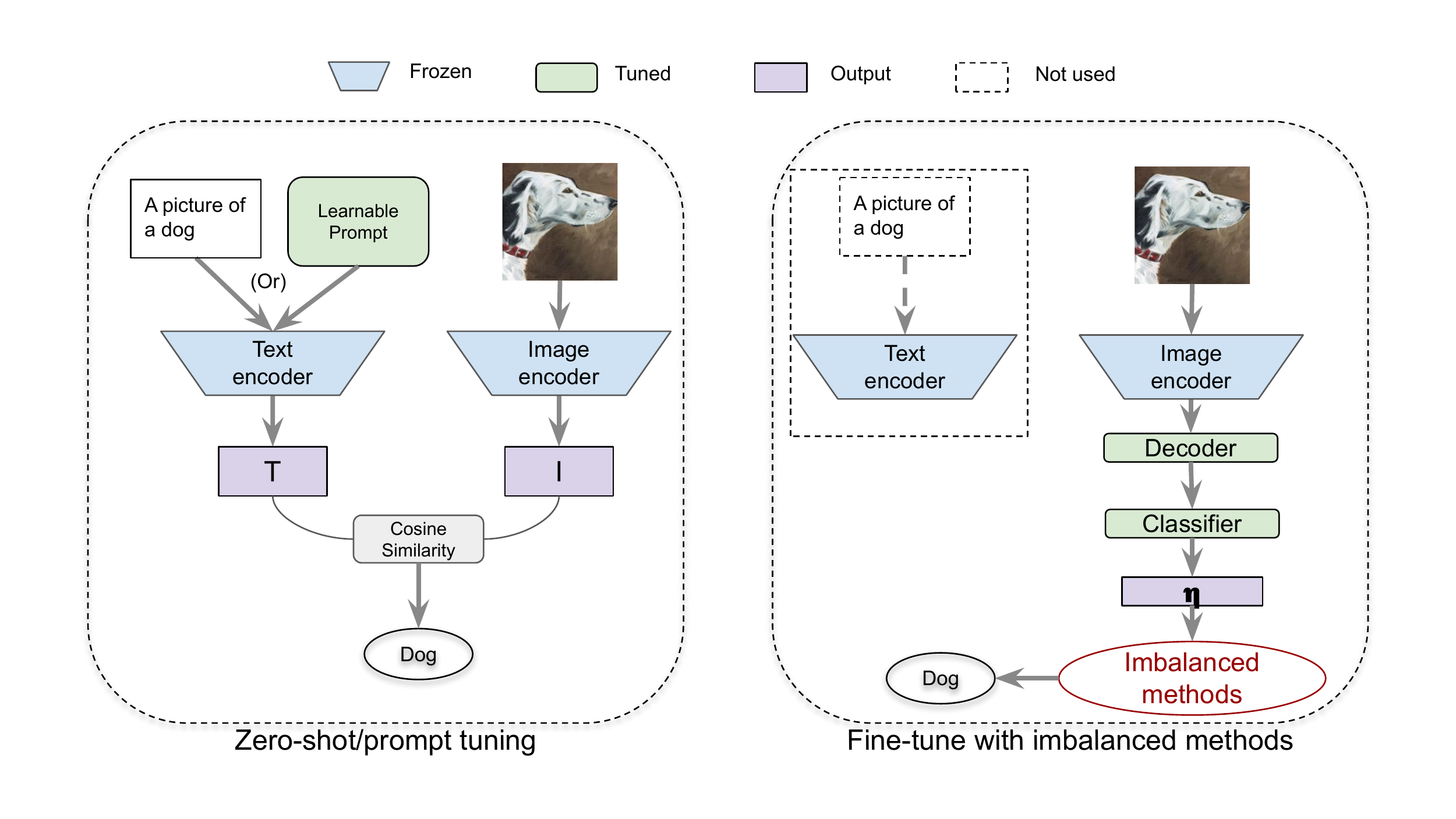}
\vspace{-.2in}
\caption{\textbf{Left:} Zero-shot and prompt tuning of VLMs for imbalanced classification. \textbf{Right:} Fine-tuning and incorporating imbalanced learning algorithms for VLMs. \textbf{Note that:} 1) We do \emph{not} use the text encoder of VLMs since adding this module makes it hard for linear probing and integrating with imbalanced methods. 2) We add a \emph{decoder} after VLMs and incorporate it with different imbalanced methods to capture nuanced features of the minority classes.}
\label{fig-main}
\end{figure*}

\subsection{Zero-shot Classification by VLMs}

In this paper, we focus on the vision-language models trained using huge image-text pairs due to their superior performance. 
During training, the contrastive language-image pre-training based methods~\citep{radford2021learning,li2022blip} train a neural network to understand the relationship between natural language and visual information. The network is trained on a large-scale dataset of image-text pairs using a contrastive learning objective. The objective is to maximize the similarity between the representations of a given image $\mathbf{I}$ and its associated text $\mathbf{T}$ while minimizing the similarity between the representations of the image and all other text $\mathbf{T'}$. 

The pre-trained VLMs can be used for zero-shot prediction on unseen downstream tasks. The left part of \figurename~\ref{fig-main} is the standard process of zero-shot prediction. To make a prediction, CLIP computes the similarity between the image representations and given text queries representations using the dot product. 
For instance, the text queries are typically natural language descriptions of the desired image classes, such as ``{\ttfamily This is a photo of [Class]}'', where ``{\ttfamily [Class]}'' can be any category in the dataset such as ``{\ttfamily cat}'' and ``{\ttfamily dog}''.
Then, VLMs return a ranking of the most likely image classes based on their similarity scores.
Formally speaking, the prediction for an input $\mathbf{x}$ is obtained as:
\begin{equation}
\label{eq-clip}
    p(y=i | \mathbf{x}) = \frac{\exp(\cos(\mathbf{T}_i, \mathbf{I}))}{\sum_{j=1}^K \exp(\cos(\mathbf{T}_j, \mathbf{I}))},
\end{equation}
where $\mathbf{T}_j$ denotes the embedding of class $j$ and $\cos(\cdot \ ,\ \cdot)$ refers to cosine similarity.

\figurename~\ref{fig-comp} shows the summarized results of zero-shot classification.
We see that VLMs can perform well on some datasets, but they may not generalize well to all datasets, especially those with different characteristics than the pre-training data.

\section{Improving VLMs for Imbalanced Learning}

Our analysis in last section demonstrates that the performance of VLMs on imbalanced tasks is not guaranteed, which motivates our further exploration: \emph{How to improve the performance of VLMs for imbalanced learning?}

\subsection{Prompt Tuning}

Prompt tuning such as CoOp~\citep{zhou2022learning} and CoCoOp~\citep{zhou2022conditional} improves the generalization of VLMs by learning its textual prompts rather than manually input as shown in the left part of \figurename~\ref{fig-main}. During training, they model prompts' context words with learnable vectors while keeping the entire pre-trained parameters fixed.
The training objective is similar to that of CLIP (Eq.~\ref{eq-clip}):
\begin{small}
\begin{equation}
\begin{split}
 &\ell(\mathcal{D}_s;\tilde{\theta}_e,\tilde{\theta}_t,\theta_p) =\\
 &-\frac{1}{s}\sum_{i=1}^s\sum_{j=1}^K \log\left(\frac{\exp(z_{ij}/\tau)}{\sum_{m=1}^K \exp(z_{im}/\tau)}\right),
\end{split}
\end{equation}
\end{small}
where $\mathcal{D}_s$ denotes an $s$-sized batch of text-image samples, $\tilde{\theta}_e$ refers to the frozen image encoder, $\tilde{\theta}_t$ refers to the frozen text encoder, $\theta_p$ is the learnable prompt, and $\tau$ is temperature.

However, prompt tuning is not designed specifically for imbalanced learning tasks since its training objective aims to calculate similarity between every image and continuous vector for all class labels.
More importantly, this method may become computationally infeasible as the number of classes scale up since the overall number of tokens to be encoded during training is $\Theta(|\mathcal{D}| K N)$, where $N$ refers to the max input length of the text encoder.\footnote{In fact, even with powerful hardware such as NVIDIA A100 (80G), it may not be feasible to run CoOp on large datasets with a huge number of classes as 8142.}

\subsection{Fine-tuning}

Other than prompt tuning, one can perform fine-tuning (linear probing or full finetuning) on training data. To conduct standard ERM~\citep{vapnik1991principles} with limited resources, we use only frozen image encoder as the representation model in linear probing as a baseline\footnote{We also provide the results of full finetuning CLIP-ViTL14. Linear probing needs single 4090 with a batch size of 256 while full finetuning requires 2 A100-80G with a batch size of 128.}. Specifically, we can replace the visual projection layer of VLMs which was used to compute the similarities between text and vision encoder with a simple fully-connected layer (classifier) for supervised training. Note that in linear probing, only the added classifier is trained while other parameters are all frozen. The training objective can be written as:
\begin{small}
\begin{equation}
    \ell(\mathcal{D}_s;\tilde{\theta}_e,\theta_c) = -\frac{1}{s} \sum_{i=1}^s \log \left(\frac{\exp (\eta_{y_i})}{\sum_{j=1}^K  \exp(\eta_j)}\right),
\end{equation}
\end{small}
where $\mathcal{D}_s$ denotes an $s$-sized batch of samples from the original training set $\mathcal{D}$, $\tilde{\theta}_e$ refers to the frozen image encoder and $\theta_c$ is the tuned classifier. However, fine-tuning the model on only the added classifier may not be sufficient to capture the nuanced features of the minority class as the VLMs themselves may lack exposure to the tail classes during pre-training. To better model the representation of tail classes, additional representation model between CLIP and classifier is necessarily needed.

\subsection{Incorporating Imbalanced Learning into VLMs}

To better model the features of tail classes and reduce the excessive memory demand, motivated by~\citep{luddecke2022image}, we freeze the VLMs that capture the general image representation, then we propose to train a light decoder to extract comprehensive image representations.

Specifically, to better \emph{adapt} VLMs on imbalanced downstream tasks, we further incorporate existing imbalanced learning algorithms into VLMs as shown in the right part of \figurename~\ref{fig-main}.
Unlike zero-shot and prompt tuning, we do not use text encoder since it is difficult to combine text encoder with other imbalanced methods.
Instead, we add a lightweight decoder after the frozen image encoder.
The output of the decoder, which is the representation of the given image will then be fed into a classifier for classification.
The classification scores (logits) $\eta$ can be easily adopted by imbalanced classification methods.
This allows us to leverage the rich knowledge in VLMs while still adapting them to specific tasks or datasets.

After the incorporation of CLIP and decoder, we provide unified formulation to existing imbalanced learning algorithms for further improvement.
Specifically, recent approaches can be broadly categorized into three types: standard training, loss function engineering, and two-stage training methods.

\textbf{Training by instance-based sampling and cross-entropy loss.} The first stage using cross-entropy loss denotes the standard training using the instance-balanced sampling and the cross entropy loss.
The training objective is:
\begin{small}
\begin{equation}
    \ell(\mathcal{D}_s;\tilde{\theta}_e,\theta_d,\theta_c) = -\frac{1}{s} \sum_{i=1}^s \log \left(\frac{\exp (\eta_{y_i})}{\sum_{j=1}^K  \exp(\eta_j)}\right),
\end{equation}
\end{small}
where $\mathcal{D}_s$ denotes an $s$-sized batch of samples from the original training set $\mathcal{D}$. This is also called Empirical risk minimization~\citep{vapnik1991principles}. \emph{Note that the representation function is composed of encoder $\tilde{\theta}_e$ and decoder $\theta_d$ and parameters with $\tilde{\cdot}$ are fixed during training.}

\textbf{Training by class-specific loss.} The first stage using adjusted cross-entropy loss denotes the
class balanced loss which adds class-specific loss weights or class-specific logit biases to obtain class balanced gradients during training~\citep{ren2020balanced,hong2021disentangling,focal,LDAM}.
The training objective is formulated as:
\begin{footnotesize}
\begin{equation}
 \ell(\mathcal{D}_s;\tilde{\theta}_e,\theta_d,\theta_c) = -\frac{1}{s} \sum_{i=1}^s \textcolor{red}{w_{y_i}} \log \left(\frac{\exp (\eta_{y_i}+\textcolor{blue}{\delta_{y_i}})}{\sum_{j=1}^K  \exp(\eta_j+\textcolor{blue}{\delta_{j}})}\right),
\end{equation}
\end{footnotesize}
where $\textcolor{red}{w_{y_i}}$ denotes the learnable weight for class $i$, and $\textcolor{blue}{\delta_{j}}$ denotes the class-specific bias for logits to obtain balanced gradients during training.

\textbf{Training by two-stage algorithms.}
Recently, two-stage algorithms~\citep{kang2019decoupling,zhang2021distribution,wang2022margin} have been proposed to overcome the issue of biased classifier in imbalanced learning. Specifically, it retrains learnable additional adjusters to calibrate the logits with fixed backbone after standard training, following the observation that standard training can obtain good representation but bad classifier.
The training objective can be summarized as:
\begin{footnotesize}
\begin{equation}
 \ell(\mathcal{D}_s;\tilde{\theta}_e,\tilde{\theta}_d,\tilde{\theta}_c,\textcolor{brown}{\theta_a}) =-\frac{1}{s} \sum_{i=1}^s \textcolor{red}{w_{y_i}} \log \left(\frac{\exp (\eta_{y_i}+\textcolor{blue}{\delta_{y_i}})}{\sum_{j=1}^K  \exp(\eta_j+\textcolor{blue}{\delta_{j}})}\right),
\end{equation}
\end{footnotesize}
where $\textcolor{brown}{\theta_a}$ denote additional trainable parameters to adjust the original biased logits for balanced predictions over classes.

The training procedure is in Algorithm~\ref{alg:two stage training}.

\section{Experiments}

We conduct extensive experiments on popular imbalanced benchmarks for exploration.
Overall, our analysis indicate that combining vision-language models with imbalanced learning is a promising approach to improve its performance.

\begin{table*}[htbp]
\centering
\caption{Results on ImageNet-LT dataset. Note that the average overall accuracy of CLIP with decoder and imbalanced algorithms is 77.12\%, representing an average improvement of \textbf{6.58}\% over the zero-shot accuracy of 70.54\%.}
\label{tab-imagenet}
 \resizebox{.95\linewidth}{!}{
\begin{tabular}{lccccccc}
\toprule
\multicolumn{1}{c|}{\multirow{2}{*}{Method}} & \multicolumn{4}{c|}{Accuracy} & \multicolumn{3}{c}{P-R-F1 score} \\  
\multicolumn{1}{c|}{} & \multicolumn{1}{c|}{Overall} & \multicolumn{1}{c|}{Many-shot} & \multicolumn{1}{c|}{Medium-shot} & \multicolumn{1}{c|}{Few-shot} & \multicolumn{1}{c|}{Precision} & \multicolumn{1}{c|}{Recall} & F1 \\ \midrule
\multicolumn{1}{l|}{Zero-shot CLIP~\citep{radford2021learning}} & \multicolumn{1}{c|}{70.54} & \multicolumn{1}{c|}{81.96} & \multicolumn{1}{c|}{70.42} & \multicolumn{1}{c|}{69.64} & \multicolumn{1}{c|}{72.13} & \multicolumn{1}{c|}{70.54} & 69.50 \\ 
\multicolumn{1}{l|}{CLIP+Linear probing} & \multicolumn{1}{c|}{67.38} & \multicolumn{1}{c|}{\underline{87.32}} & \multicolumn{1}{c|}{65.09} & \multicolumn{1}{c|}{18.97} & \multicolumn{1}{c|}{69.76} & \multicolumn{1}{c|}{67.38} & 63.88 \\

\multicolumn{1}{l|}{CLIP+Full finetuning} & \multicolumn{1}{c|}{68.54} & \multicolumn{1}{c|}{83.03} & \multicolumn{1}{c|}{65.03} & \multicolumn{1}{c|}{39.87} & \multicolumn{1}{c|}{74.43} & \multicolumn{1}{c|}{68.54} & 67.19 \\ 

\multicolumn{1}{l|}{CoOp~\citep{zhou2022learning}} & \multicolumn{1}{c|}{60.30} & \multicolumn{1}{c|}{-} & \multicolumn{1}{c|}{-} & \multicolumn{1}{c|}{-} & \multicolumn{1}{c|}{-} & \multicolumn{1}{c|}{-} & 59.00 \\ \midrule
\multicolumn{8}{c}{CLIP + imbalanced learning algorithms} \\ \midrule
\multicolumn{1}{l|}{Softmax} & \multicolumn{1}{c|}{75.05} & \multicolumn{1}{c|}{87.18} & \multicolumn{1}{c|}{72.67} & \multicolumn{1}{c|}{49.10} & \multicolumn{1}{c|}{78.56} & \multicolumn{1}{c|}{75.05} & 73.96 \\ 
\multicolumn{1}{l|}{CBW} & \multicolumn{1}{c|}{77.37} & \multicolumn{1}{c|}{85.44} & \multicolumn{1}{c|}{76.19} & \multicolumn{1}{c|}{58.64} & \multicolumn{1}{c|}{79.03} & \multicolumn{1}{c|}{77.37} & 76.59 \\ 
\multicolumn{1}{l|}{Focal Loss~\citep{focal}} & \multicolumn{1}{c|}{74.27} & \multicolumn{1}{c|}{86.66} & \multicolumn{1}{c|}{71.87} & \multicolumn{1}{c|}{47.63} & \multicolumn{1}{c|}{78.25} & \multicolumn{1}{c|}{74.27} & 73.26 \\ 
\multicolumn{1}{l|}{LDAM Loss~\citep{LDAM}} & \multicolumn{1}{c|}{75.29} & \multicolumn{1}{c|}{87.10} & \multicolumn{1}{c|}{72.98} & \multicolumn{1}{c|}{50.01} & \multicolumn{1}{c|}{78.73} & \multicolumn{1}{c|}{75.29} & 74.26 \\ 
\multicolumn{1}{l|}{Balanced Softmax~\citep{ren2020balanced}} & \multicolumn{1}{c|}{\underline{79.03}} & \multicolumn{1}{c|}{83.61} & \multicolumn{1}{c|}{77.82} & \multicolumn{1}{c|}{\underline{70.37}} & \multicolumn{1}{c|}{79.70} & \multicolumn{1}{c|}{79.03} & 78.69 \\ 
\multicolumn{1}{l|}{LADE Loss~\citep{hong2021disentangling}} & \multicolumn{1}{c|}{\textbf{79.44}} & \multicolumn{1}{c|}{83.60} & \multicolumn{1}{c|}{78.14} & \multicolumn{1}{c|}{\textbf{72.22}} & \multicolumn{1}{c|}{\textbf{80.06}} & \multicolumn{1}{c|}{\textbf{79.44}} & \textbf{79.13} \\ 
\multicolumn{1}{l|}{CRT~\citep{kang2019decoupling}} & \multicolumn{1}{c|}{75.22} & \multicolumn{1}{c|}{\textbf{87.55}} & \multicolumn{1}{c|}{72.78} & \multicolumn{1}{c|}{48.95} & \multicolumn{1}{c|}{78.33} & \multicolumn{1}{c|}{75.22} & 74.05 \\ 
\multicolumn{1}{l|}{LWS~\citep{kang2019decoupling}} & \multicolumn{1}{c|}{76.98} & \multicolumn{1}{c|}{87.08} & \multicolumn{1}{c|}{75.45} & \multicolumn{1}{c|}{53.78} & \multicolumn{1}{c|}{79.22} & \multicolumn{1}{c|}{76.98} & 76.09 \\ 
\multicolumn{1}{l|}{Disalign~\citep{zhang2021distribution}} & \multicolumn{1}{c|}{79.26} & \multicolumn{1}{c|}{83.56} & \multicolumn{1}{c|}{\textbf{78.38}} & \multicolumn{1}{c|}{70.18} & \multicolumn{1}{c|}{79.82} & \multicolumn{1}{c|}{79.26} & 78.94 \\ 
\multicolumn{1}{l|}{MARC~\citep{wang2022margin}} & \multicolumn{1}{c|}{79.32} & \multicolumn{1}{c|}{83.91} & \multicolumn{1}{c|}{\underline{78.28}} & \multicolumn{1}{c|}{70.02} & \multicolumn{1}{c|}{\underline{79.92}} & \multicolumn{1}{c|}{\underline{79.32}} & \underline{79.01} \\ \bottomrule
\end{tabular}
}
\end{table*}

\begin{table*}[htbp]
\centering
\caption{Results on iNaturalist dataset. Note that the average overall accuracy of CLIP with decoder and imbalanced algorithms is 69.82\%, representing an average improvement of \textbf{64.37}\% over the zero-shot accuracy of 5.45\%.}
\label{tab-inat}
 \resizebox{.95\linewidth}{!}{

\begin{tabular}{lccccccc}
\toprule
\multicolumn{1}{c|}{\multirow{2}{*}{Method}} &
  \multicolumn{4}{c|}{Accuracy} &
  \multicolumn{3}{c}{P-R-F1 score} \\
\multicolumn{1}{c|}{} &
  \multicolumn{1}{c|}{Overall} &
  \multicolumn{1}{c|}{Many-shot} &
  \multicolumn{1}{c|}{Medium-shot} &
  \multicolumn{1}{c|}{Few-shot} &
  \multicolumn{1}{c|}{Precision} &
  \multicolumn{1}{c|}{Recall} &
  F1 \\ \midrule
\multicolumn{1}{l|}{Zero-shot CLIP~\citep{radford2021learning}} &
  \multicolumn{1}{c|}{5.45} &
  \multicolumn{1}{c|}{9.87} &
  \multicolumn{1}{c|}{5.28} &
  \multicolumn{1}{c|}{4.59} &
  \multicolumn{1}{c|}{3.85} &
  \multicolumn{1}{c|}{5.45} &
  3.70 \\
\multicolumn{1}{l|}{CLIP+Linear probing} &
  \multicolumn{1}{c|}{10.03} &
  \multicolumn{1}{c|}{62.35} &
  \multicolumn{1}{c|}{7.10} &
  \multicolumn{1}{c|}{0.07} &
  \multicolumn{1}{c|}{4.54} &
  \multicolumn{1}{c|}{10.03} &
  4.78 \\

\multicolumn{1}{l|}{CLIP+Full finetuning} &
  \multicolumn{1}{c|}{65.43} &
  \multicolumn{1}{c|}{\textbf{79.41}} &
  \multicolumn{1}{c|}{67.58} &
  \multicolumn{1}{c|}{59.07} &
  \multicolumn{1}{c|}{70.82} &
  \multicolumn{1}{c|}{65.43} &
  63.80 \\
  
\multicolumn{1}{l|}{CoOp~\citep{zhou2022learning}} &
  \multicolumn{1}{c|}{-} &
  \multicolumn{1}{c|}{-} &
  \multicolumn{1}{c|}{-} &
  \multicolumn{1}{c|}{-} &
  \multicolumn{1}{c|}{-} &
  \multicolumn{1}{c|}{-} &
  - \\ \midrule
\multicolumn{8}{c}{CLIP + imbalanced learning algorithms} \\ \midrule
\multicolumn{1}{l|}{Softmax} &
  \multicolumn{1}{c|}{65.57} &
  \multicolumn{1}{c|}{{76.54}} &
  \multicolumn{1}{c|}{68.31} &
  \multicolumn{1}{c|}{59.25} &
  \multicolumn{1}{c|}{70.76} &
  \multicolumn{1}{c|}{65.57} &
  64.15 \\
\multicolumn{1}{l|}{CBW} &
  \multicolumn{1}{c|}{70.33} &
  \multicolumn{1}{c|}{65.56} &
  \multicolumn{1}{c|}{71.59} &
  \multicolumn{1}{c|}{69.99} &
  \multicolumn{1}{c|}{73.83} &
  \multicolumn{1}{c|}{70.33} &
  68.98 \\
\multicolumn{1}{l|}{Focal Loss~\citep{focal}} &
  \multicolumn{1}{c|}{64.81} &
  \multicolumn{1}{c|}{75.81} &
  \multicolumn{1}{c|}{67.65} &
  \multicolumn{1}{c|}{58.36} &
  \multicolumn{1}{c|}{70.44} &
  \multicolumn{1}{c|}{64.81} &
  63.47 \\
\multicolumn{1}{l|}{LDAM Loss~\citep{LDAM}} &
  \multicolumn{1}{c|}{66.02} &
  \multicolumn{1}{c|}{\underline{76.68}} &
  \multicolumn{1}{c|}{68.53} &
  \multicolumn{1}{c|}{60.06} &
  \multicolumn{1}{c|}{71.13} &
  \multicolumn{1}{c|}{66.02} &
  64.61 \\
\multicolumn{1}{l|}{Balanced Softmax~\citep{ren2020balanced}} &
  \multicolumn{1}{c|}{70.59} &
  \multicolumn{1}{c|}{68.43} &
  \multicolumn{1}{c|}{71.30} &
  \multicolumn{1}{c|}{70.25} &
  \multicolumn{1}{c|}{73.87} &
  \multicolumn{1}{c|}{70.59} &
  69.20 \\
\multicolumn{1}{l|}{LADE Loss~\citep{hong2021disentangling}} &
  \multicolumn{1}{c|}{70.90} &
  \multicolumn{1}{c|}{67.96} &
  \multicolumn{1}{c|}{71.52} &
  \multicolumn{1}{c|}{70.89} &
  \multicolumn{1}{c|}{74.16} &
  \multicolumn{1}{c|}{70.90} &
  69.54 \\
\multicolumn{1}{l|}{CRT~\citep{kang2019decoupling}} &
  \multicolumn{1}{c|}{\textbf{73.24}} &
  \multicolumn{1}{c|}{72.18} &
  \multicolumn{1}{c|}{\textbf{74.36}} &
  \multicolumn{1}{c|}{72.10} &
  \multicolumn{1}{c|}{\textbf{76.87}} &
  \multicolumn{1}{c|}{\textbf{73.24}} &
  \textbf{72.22} \\
\multicolumn{1}{l|}{LWS~\citep{kang2019decoupling}} &
  \multicolumn{1}{c|}{{\underline{72.63}}} &
  \multicolumn{1}{c|}{70.37} &
  \multicolumn{1}{c|}{{\underline{73.82}}} &
  \multicolumn{1}{c|}{71.73} &
  \multicolumn{1}{c|}{{\underline{75.52}}} &
  \multicolumn{1}{c|}{{\underline{72.63}}} &
  {\underline{71.54}} \\
\multicolumn{1}{l|}{Disalign~\citep{zhang2021distribution}} &
  \multicolumn{1}{c|}{72.33} &
  \multicolumn{1}{c|}{65.46} &
  \multicolumn{1}{c|}{73.20} &
  \multicolumn{1}{c|}{\textbf{73.02}} &
  \multicolumn{1}{c|}{75.14} &
  \multicolumn{1}{c|}{72.33} &
  71.14 \\
\multicolumn{1}{l|}{MARC~\citep{wang2022margin}} &
  \multicolumn{1}{c|}{71.82} &
  \multicolumn{1}{c|}{64.87} &
  \multicolumn{1}{c|}{72.64} &
  \multicolumn{1}{c|}{{\underline{72.59}}} &
  \multicolumn{1}{c|}{74.89} &
  \multicolumn{1}{c|}{71.82} &
  70.56 \\ \bottomrule
\end{tabular}
 
}
\end{table*}

\begin{table*}[htbp]
\centering
\caption{Results on Places-LT dataset. Note that the average overall accuracy of CLIP with decoder and imbalanced algorithms is 43.86\%, representing an average improvement of \textbf{6.17}\% over the zero-shot accuracy of 37.69\%.}
\label{tab-places}
 \resizebox{.95\linewidth}{!}{

 \begin{tabular}{lccccccc}
\toprule
\multicolumn{1}{c|}{\multirow{2}{*}{Method}} &
  \multicolumn{4}{c|}{Accuracy} &
  \multicolumn{3}{c}{P-R-F1 score} \\
\multicolumn{1}{c|}{} &
  \multicolumn{1}{c|}{Overall} &
  \multicolumn{1}{c|}{Many-shot} &
  \multicolumn{1}{c|}{Medium-shot} &
  \multicolumn{1}{c|}{Few-shot} &
  \multicolumn{1}{c|}{Precision} &
  \multicolumn{1}{c|}{Recall} &
  F1 \\ \midrule
\multicolumn{1}{l|}{Zero-shot CLIP~\citep{radford2021learning}} &
  \multicolumn{1}{c|}{37.69} &
  \multicolumn{1}{c|}{40.94} &
  \multicolumn{1}{c|}{35.70} &
  \multicolumn{1}{c|}{44.64} &
  \multicolumn{1}{c|}{39.25} &
  \multicolumn{1}{c|}{37.69} &
  36.52 \\
\multicolumn{1}{l|}{CLIP+Linear probing} &
  \multicolumn{1}{c|}{38.18} &
  \multicolumn{1}{c|}{\textbf{55.69}} &
  \multicolumn{1}{c|}{34.47} &
  \multicolumn{1}{c|}{14.41} &
  \multicolumn{1}{c|}{48.53} &
  \multicolumn{1}{c|}{38.18} &
  36.10 \\

\multicolumn{1}{l|}{CLIP+Full finetuning} &
  \multicolumn{1}{c|}{38.24} &
  \multicolumn{1}{c|}{52.82} &
  \multicolumn{1}{c|}{34.55} &
  \multicolumn{1}{c|}{19.82} &
  \multicolumn{1}{c|}{48.94} &
  \multicolumn{1}{c|}{38.24} &
 37.00 \\
  
\multicolumn{1}{l|}{CoOp~\citep{zhou2022learning}} &
  \multicolumn{1}{c|}{26.10} &
  \multicolumn{1}{c|}{-} &
  \multicolumn{1}{c|}{-} &
  \multicolumn{1}{c|}{-} &
  \multicolumn{1}{c|}{-} &
  \multicolumn{1}{c|}{-} &
  24.00 \\ \midrule
\multicolumn{8}{c}{CLIP + imbalanced learning algorithms} \\ \midrule
\multicolumn{1}{l|}{Softmax} &
  \multicolumn{1}{c|}{40.15} &
  \multicolumn{1}{c|}{53.75} &
  \multicolumn{1}{c|}{36.48} &
  \multicolumn{1}{c|}{23.47} &
  \multicolumn{1}{c|}{49.93} &
  \multicolumn{1}{c|}{40.15} &
  38.60 \\
\multicolumn{1}{l|}{CBW} &
  \multicolumn{1}{c|}{44.94} &
  \multicolumn{1}{c|}{49.55} &
  \multicolumn{1}{c|}{45.61} &
  \multicolumn{1}{c|}{34.88} &
  \multicolumn{1}{c|}{47.37} &
  \multicolumn{1}{c|}{44.94} &
  43.81 \\
\multicolumn{1}{l|}{Focal Loss~\citep{focal}} &
  \multicolumn{1}{c|}{39.83} &
  \multicolumn{1}{c|}{52.87} &
  \multicolumn{1}{c|}{36.26} &
  \multicolumn{1}{c|}{23.94} &
  \multicolumn{1}{c|}{49.93} &
  \multicolumn{1}{c|}{39.83} &
  38.61 \\
\multicolumn{1}{l|}{LDAM Loss~\citep{LDAM}} &
  \multicolumn{1}{c|}{40.34} &
  \multicolumn{1}{c|}{{\underline {54.60}}} &
  \multicolumn{1}{c|}{36.57} &
  \multicolumn{1}{c|}{22.7} &
  \multicolumn{1}{c|}{50.17} &
  \multicolumn{1}{c|}{40.34} &
  38.76 \\
\multicolumn{1}{l|}{Balanced Softmax~\citep{ren2020balanced}} &
  \multicolumn{1}{c|}{47.36} &
  \multicolumn{1}{c|}{50.18} &
  \multicolumn{1}{c|}{47.10} &
  \multicolumn{1}{c|}{42.76} &
  \multicolumn{1}{c|}{49.52} &
  \multicolumn{1}{c|}{47.36} &
  46.42 \\
\multicolumn{1}{l|}{LADE Loss~\citep{hong2021disentangling}} &
  \multicolumn{1}{c|}{47.29} &
  \multicolumn{1}{c|}{50.03} &
  \multicolumn{1}{c|}{46.84} &
  \multicolumn{1}{c|}{43.26} &
  \multicolumn{1}{c|}{49.19} &
  \multicolumn{1}{c|}{47.29} &
  46.32 \\
\multicolumn{1}{l|}{CRT~\citep{kang2019decoupling}} &
  \multicolumn{1}{c|}{41.91} &
  \multicolumn{1}{c|}{53.31} &
  \multicolumn{1}{c|}{38.16} &
  \multicolumn{1}{c|}{29.48} &
  \multicolumn{1}{c|}{{\underline{50.75}}} &
  \multicolumn{1}{c|}{41.91} &
  40.67 \\
\multicolumn{1}{l|}{LWS~\citep{kang2019decoupling}} &
  \multicolumn{1}{c|}{39.72} &
  \multicolumn{1}{c|}{51.47} &
  \multicolumn{1}{c|}{35.30} &
  \multicolumn{1}{c|}{28.21} &
  \multicolumn{1}{c|}{\textbf{51.14}} &
  \multicolumn{1}{c|}{39.72} &
  39.33 \\
\multicolumn{1}{l|}{Disalign~\citep{zhang2021distribution}} &
  \multicolumn{1}{c|}{{\underline {48.43}}} &
  \multicolumn{1}{c|}{47.69} &
  \multicolumn{1}{c|}{{\underline {49.92}}} &
  \multicolumn{1}{c|}{{\underline {46.37}}} &
  \multicolumn{1}{c|}{48.54} &
  \multicolumn{1}{c|}{{\underline {48.43}}} &
  {\underline {47.94}} \\
\multicolumn{1}{l|}{MARC~\citep{wang2022margin}} &
  \multicolumn{1}{c|}{\textbf{48.66}} &
  \multicolumn{1}{c|}{47.43} &
  \multicolumn{1}{c|}{\textbf{50.02}} &
  \multicolumn{1}{c|}{\textbf{47.81}} &
  \multicolumn{1}{c|}{48.72} &
  \multicolumn{1}{c|}{\textbf{48.66}} &
  \textbf{48.15} \\ 
  \bottomrule
\end{tabular}
 
}
\end{table*}

\subsection{Setup}

\subsubsection{Datasets}
We adopt the standard evaluation protocol~\citep{liu2019large} on three commonly used imbalanced benchmarks: ImageNet-LT~\citep{liu2019large}, Places-LT~\citep{zhou2017places}, and iNaturalist2018~\citep{van2018inaturalist}.
The training data class distributions of these datasets are shown in \figurename~\ref{fig-dist}. Note that the test sets are balanced to ensure the fairness on all classes during evaluation.
The detailed dataset statistics are in Appendix~\ref{sec-append-data}.

For rigorous evaluation, we split the classes in each benchmark into three groups based on the number of images available for each class. These groups are known as Many-shot (with more than 100 images), Medium-shot (with 20 to 100 images), and Few-shot (with less than 20 images). We then evaluate the performance of the selected models across a range of difficulty levels, from classes with abundant training data to those with very few examples.

\subsubsection{Models, baselines, and metrics}
We adopt CLIP~\citep{radford2021learning} as the main VLM for experiments.
Specifically, we use the ViT-L~\citep{dosovitskiyimage} backbone for CLIP and more ablations on backbones are in Sec.~\ref{sec-exp-backbone}.
We select three baselines: 1) zero-shot prediction by VLMs as introduced in Sec.~\ref{sec-zeroshot}; 2) fine-tuning (full finetuning \& linear probing) VLMs; and 3) prompt tuning (CoOp~\citep{zhou2022learning}).
These baselines are compared with the further incorporation of existing imbalanced learning approaches to VLMs.
Appendix~\ref{intro-algs} introduces the used imbalanced learning algorithms.

We use two metrics for thorough evaluation: 1) accuracy, including overall, many-shot, medium-shot, and few-shot accuracy; 2) P-R-F1 score, including precision, recall, and F1 score.

\begin{algorithm*}[htbp]
\caption{The torch-like code for creating the decoder.}
\label{alg:decoder}
\begin{algorithmic}[1]

\State{from torch import nn}
\State{from timm.models.vision\_transformer import Block}
\State{decoder\_blocks = nn.Sequential(*[Block(hidden\_size, decoder\_num\_heads, mlp\_ratio, qkv\_bias=True) for i in range(decoder\_depth)])}
\end{algorithmic}
\end{algorithm*}

\subsubsection{Training details}
\label{exp-train}

The decoder in \figurename~\ref{fig-main} comprises three transformer blocks, where each block constitutes a multi-head attention mechanism and a feed-forward neural network. The former, with four attention heads, facilitates an enriched modeling of extensive range dependencies inherent in the input sequence. Meanwhile, the latter comprises a multi-layer perceptron (MLP) with a dropout ratio set at 0.5. As Algorithm~\ref{alg:decoder} shows, each Transformer block can be adopted from the PyTorch Image Models (TIMM) library's vision\_transformer module, with specified parameters relating to the hidden size, number of attention heads, and the MLP ratio. This construction ensures a robust and efficient transformation of the encoded inputs into the desired outputs. The training process involves 8192 iterations with a batch size of 256, using the SGD optimizer with momentum 0.9 and weight decay of $5e-4$ for all datasets. The learning rate schedule follows a cosine function with an initial value of 0.03, which gradually decays to 0. In addition, a warm-up strategy is employed for the learning rate scheduler, with warm-up iterations of 512. Note that for Full finetuning, to get better results and avoid OOM, we use a batch size of 128, the AdamW optimizer with a learning rate of 0.00003. More detailed information on the hyperparameters and training logs are at \url{https://github.com/Imbalance-LVM/Imbalance-LVM}.

\begin{table*}[t!]
\centering
\caption{Comparisons between ViT-B16 and ViT-L14 variants of CLIP backbone on all datasets.}
\label{tab-backbones}
 \resizebox{.5\textwidth}{!}{
 \begin{tabular}{c|c|c|c|c}
\toprule
Method                            & Backbone & ImageNet-LT & Places-LT & iNaturalist  \\ \midrule
\multirow{2}{*}{One-stage} & ViT-B16  & 68.14            & 37.4           & 51.35            \\
                                  & ViT-L14  & \textbf{75.05}   & \textbf{40.15} & \textbf{65.57}   \\ \midrule
\multirow{2}{*}{Two-stage} & ViT-B16  & 73.23            & 46.78          & 59.22            \\
                                  & ViT-L14  & \textbf{79.26}   & \textbf{48.43} & \textbf{72.33}   \\ \bottomrule
\end{tabular}
}
\end{table*}

\begin{table*}[t!]
\caption{Comparisons between ViT of CLIP (400M) and Laion-CLIP (2B) on iNaturalist18 and Places-LT.} 
\label{tab-laion}
\resizebox{.98\textwidth}{!}{
\begin{tabular}{c|c|c|cccc|ccc}
\toprule
 \multirow{2}{*}{Method} & \multirow{2}{*}{Dataset} & \multirow{2}{*}{Ablation} & \multicolumn{4}{c|}{Accuracy} & \multicolumn{3}{c}{P-R-F1 score} \\  
 & & & \multicolumn{1}{c|}{Overall} & \multicolumn{1}{c|}{Many-shot} & \multicolumn{1}{c|}{Medium-shot} & Few-shot & \multicolumn{1}{c|}{Precision} & \multicolumn{1}{c|}{Recall} & F1 \\ \midrule
 \multirow{4}{*}{Zero-shot} & \multirow{2}{*}{iNaturalist18} & Laion-CLIP & \multicolumn{1}{c|}{3.82} & \multicolumn{1}{c|}{6.34} & \multicolumn{1}{c|}{3.57} & 3.38 & \multicolumn{1}{c|}{2.18} & \multicolumn{1}{c|}{3.81} & 2.26 \\ 
 & & CLIP & \multicolumn{1}{c|}{\textbf{5.45}} & \multicolumn{1}{c|}{\textbf{9.87}} & \multicolumn{1}{c|}{\textbf{5.28}} & \textbf{4.59} & \multicolumn{1}{c|}{\textbf{3.85}} & \multicolumn{1}{c|}{\textbf{5.45}} & \textbf{3.70} \\ 
 & \multirow{2}{*}{Places-LT} & Laion-CLIP & \multicolumn{1}{c|}{\textbf{40.64}} & \multicolumn{1}{c|}{\textbf{49.31}} & \multicolumn{1}{c|}{\textbf{39.43}} & 43.41 & \multicolumn{1}{c|}{\textbf{42.57}} & \multicolumn{1}{c|}{\textbf{40.63}} & \textbf{39.71} \\ 
 & & CLIP & \multicolumn{1}{c|}{37.69} & \multicolumn{1}{c|}{40.94} & \multicolumn{1}{c|}{35.70} & \textbf{44.64} & \multicolumn{1}{c|}{39.25} & \multicolumn{1}{c|}{37.69} & 36.52 \\ \midrule
\multirow{4}{*}{Balanced SoftMax} & \multirow{2}{*}{iNaturalist18} & Laion-CLIP & \multicolumn{1}{c|}{60.94} & \multicolumn{1}{c|}{57.84} & \multicolumn{1}{c|}{60.88} & 61.82 & \multicolumn{1}{c|}{64.04} & \multicolumn{1}{c|}{60.94} & 59.20 \\ 
 & & CLIP & \multicolumn{1}{c|}{\textbf{70.59}} & \multicolumn{1}{c|}{\textbf{68.43}} & \multicolumn{1}{c|}{\textbf{71.30}} & \textbf{70.25} & \multicolumn{1}{c|}{\textbf{73.87}} & \multicolumn{1}{c|}{\textbf{70.59}} & \textbf{69.20} \\ 
  & \multirow{2}{*}{Places-LT} & Laion-CLIP & \multicolumn{1}{c|}{\textbf{47.45}} & \multicolumn{1}{c|}{48.70} & \multicolumn{1}{c|}{\textbf{48.06}} & \textbf{43.77} & \multicolumn{1}{c|}{\textbf{49.64}} & \multicolumn{1}{c|}{\textbf{47.45}} & \textbf{46.58} \\ 
 & & CLIP & \multicolumn{1}{c|}{47.36} & \multicolumn{1}{c|}{\textbf{50.18}} & \multicolumn{1}{c|}{47.10} & 42.76 & \multicolumn{1}{c|}{49.52} & \multicolumn{1}{c|}{47.36} & 46.42 \\  \midrule
 \multirow{4}{*}{DisAlign} & \multirow{2}{*}{iNaturalist18} & Laion-CLIP & \multicolumn{1}{c|}{62.61} & \multicolumn{1}{c|}{53.64} & \multicolumn{1}{c|}{62.99} & 64.46 & \multicolumn{1}{c|}{65.50} & \multicolumn{1}{c|}{62.61} & 61.20 \\ 
 & & CLIP & \multicolumn{1}{c|}{\textbf{72.33}} & \multicolumn{1}{c|}{\textbf{65.46}} & \multicolumn{1}{c|}{\textbf{73.20}} & \textbf{73.02} & \multicolumn{1}{c|}{\textbf{75.14}} & \multicolumn{1}{c|}{\textbf{72.33}} & \textbf{71.14}  \\  
 & \multirow{2}{*}{Places-LT} & Laion-CLIP & \multicolumn{1}{c|}{45.35} & \multicolumn{1}{c|}{46.26} & \multicolumn{1}{c|}{46.09} & 41.99 & \multicolumn{1}{c|}{46.80} & \multicolumn{1}{c|}{45.35} & 44.89 \\ 
  & & CLIP & \multicolumn{1}{c|}{\textbf{48.43}} & \multicolumn{1}{c|}{\textbf{47.69}} & \multicolumn{1}{c|}{\textbf{49.92}} & \textbf{46.37} & \multicolumn{1}{c|}{\textbf{48.54}} & \multicolumn{1}{c|}{\textbf{48.43}} & \textbf{47.94} \\ 
 \midrule
 \multirow{4}{*}{MARC} & \multirow{2}{*}{iNaturalist18} & Laion-CLIP & \multicolumn{1}{c|}{61.77} & \multicolumn{1}{c|}{51.35} & \multicolumn{1}{c|}{62.16} & 63.99 & \multicolumn{1}{c|}{64.88} & \multicolumn{1}{c|}{61.77} & 60.16 \\ 
 & & CLIP & \multicolumn{1}{c|}{\textbf{71.82}} & \multicolumn{1}{c|}{\textbf{64.87}} & \multicolumn{1}{c|}{\textbf{72.64}} & \textbf{72.59} & \multicolumn{1}{c|}{\textbf{74.89}} & \multicolumn{1}{c|}{\textbf{71.82}} & \textbf{70.56} \\ 
 & \multirow{2}{*}{Places-LT} & Laion-CLIP & \multicolumn{1}{c|}{45.57} & \multicolumn{1}{c|}{46.74} & \multicolumn{1}{c|}{46.09} & 42.24 & \multicolumn{1}{c|}{47.21} & \multicolumn{1}{c|}{45.57} & 45.26 \\ 
 & & CLIP & \multicolumn{1}{c|}{\textbf{48.66}} & \multicolumn{1}{c|}{\textbf{47.43}} & \multicolumn{1}{c|}{\textbf{50.02}} & \textbf{47.81} & \multicolumn{1}{c|}{\textbf{48.72}} & \multicolumn{1}{c|}{\textbf{48.66}} & \textbf{48.15} \\ 
 \bottomrule
\end{tabular}
}
\end{table*}

\begin{table*}[t!]
\caption{Results using CLIP-ResNet101 on ImageNet-LT, iNaturalist and Places-LT datasets.} 
\label{tab-resnet}
\resizebox{.98\textwidth}{!}{
\begin{tabular}{c|c|cccc|ccc}
\toprule
 \multirow{2}{*}{Dataset} & \multirow{2}{*}{Method}  & \multicolumn{4}{c|}{Accuracy} & \multicolumn{3}{c}{P-R-F1 score} \\  
 & & \multicolumn{1}{c|}{Overall} & \multicolumn{1}{c|}{Many-shot} & \multicolumn{1}{c|}{Medium-shot} & Few-shot & \multicolumn{1}{c|}{Precision} & \multicolumn{1}{c|}{Recall} & F1 \\ \midrule
 \multirow{10}{*}{ImageNet-LT}  & Zero-shot & \multicolumn{1}{c|}{53.62} & \multicolumn{1}{c|}{59.57} & \multicolumn{1}{c|}{53.57} & 52.81 & \multicolumn{1}{c|}{56.22} & \multicolumn{1}{c|}{53.62} & 52.50 \\ 
  & Linear probing & \multicolumn{1}{c|}{9.33} & \multicolumn{1}{c|}{24.23} & \multicolumn{1}{c|}{0.00} & 0.00 & \multicolumn{1}{c|}{7.66} & \multicolumn{1}{c|}{9.33} & 4.97 \\ 
  & Full finetuning & \multicolumn{1}{c|}{57.61} & \multicolumn{1}{c|}{74.49} & \multicolumn{1}{c|}{52.82} & 26.66 & \multicolumn{1}{c|}{62.89} & \multicolumn{1}{c|}{57.61} & 55.86 \\ 
  & Decoder+SoftMax & \multicolumn{1}{c|}{48.01} & \multicolumn{1}{c|}{66.93} & \multicolumn{1}{c|}{42.09} & 15.32 & \multicolumn{1}{c|}{56.75} & \multicolumn{1}{c|}{48.01} & 45.21 \\
  & Decoder+Balanced SoftMax & \multicolumn{1}{c|}{ 53.83} & \multicolumn{1}{c|}{60.60} & \multicolumn{1}{c|}{52.17} & 40.53 & \multicolumn{1}{c|}{56.01} & \multicolumn{1}{c|}{53.83} & 52.57 \\ 
  & Decoder+MARC & \multicolumn{1}{c|}{55.04} & \multicolumn{1}{c|}{58.29} & \multicolumn{1}{c|}{54.73} & \textbf{46.91} & \multicolumn{1}{c|}{56.48} & \multicolumn{1}{c|}{55.04} & 54.35 \\
  & Decoder+CRT & \multicolumn{1}{c|}{53.89} & \multicolumn{1}{c|}{66.89} & \multicolumn{1}{c|}{51.98} & 23.82 & \multicolumn{1}{c|}{57.34} & \multicolumn{1}{c|}{53.89} & 51.77 \\
  & Full finetuning+Balanced SoftMax & \multicolumn{1}{c|}{60.47} & \multicolumn{1}{c|}{69.18} & \multicolumn{1}{c|}{58.25} & 43.63 & \multicolumn{1}{c|}{62.05} & \multicolumn{1}{c|}{60.47} & \textbf{59.79} \\ 
  & Full finetuning+MARC & \multicolumn{1}{c|}{\textbf{60.97}} & \multicolumn{1}{c|}{73.73} & \multicolumn{1}{c|}{\textbf{58.69}} & 32.91 & \multicolumn{1}{c|}{63.00} & \multicolumn{1}{c|}{\textbf{60.97}} & 59.61 \\
  & Full finetuning+CRT & \multicolumn{1}{c|}{59.90} & \multicolumn{1}{c|}{\textbf{75.69}} & \multicolumn{1}{c|}{56.43} & 27.47 & \multicolumn{1}{c|}{\textbf{63.78}} & \multicolumn{1}{c|}{59.90} & 58.21 \\
  \midrule

   \multirow{10}{*}{iNaturalist}  & Zero-shot & \multicolumn{1}{c|}{2.03} & \multicolumn{1}{c|}{2.76} & \multicolumn{1}{c|}{2.00} & 1.72 & \multicolumn{1}{c|}{11.55} & \multicolumn{1}{c|}{20.27} & 11.70 \\ 
  & Linear probing & \multicolumn{1}{c|}{0.24} & \multicolumn{1}{c|}{2.34} & \multicolumn{1}{c|}{0.00} & 0.00 & \multicolumn{1}{c|}{0.01} & \multicolumn{1}{c|}{0.24} & 0.02 \\ 
  & Full finetuning & \multicolumn{1}{c|}{46.84} & \multicolumn{1}{c|}{\textbf{68.17}} & \multicolumn{1}{c|}{49.48} & 37.92 & \multicolumn{1}{c|}{50.80} & \multicolumn{1}{c|}{46.84} & 44.28 \\ 
  & Decoder+SoftMax & \multicolumn{1}{c|}{14.95} & \multicolumn{1}{c|}{28.42} & \multicolumn{1}{c|}{15.80} & 10.36 & \multicolumn{1}{c|}{16.55} & \multicolumn{1}{c|}{14.95} & 13.04 \\
  & Decoder+Balanced SoftMax & \multicolumn{1}{c|}{  21.53} & \multicolumn{1}{c|}{20.47} & \multicolumn{1}{c|}{21.79} & 21.47 & \multicolumn{1}{c|}{21.44} & \multicolumn{1}{c|}{21.53} & 18.55 \\ 
  & Decoder+MARC & \multicolumn{1}{c|}{25.14} & \multicolumn{1}{c|}{12.83} & \multicolumn{1}{c|}{25.96} & 27.31 & \multicolumn{1}{c|}{24.27} & \multicolumn{1}{c|}{25.14} & 22.08 \\
  & Decoder+CRT & \multicolumn{1}{c|}{12.77} & \multicolumn{1}{c|}{33.14} & \multicolumn{1}{c|}{14.70} & 5.01 & \multicolumn{1}{c|}{14.29} & \multicolumn{1}{c|}{12.77} & 10.78 \\
  & Full finetuning+Balanced SoftMax & \multicolumn{1}{c|}{49.36} & \multicolumn{1}{c|}{56.18} & \multicolumn{1}{c|}{50.32} & 46.37 & \multicolumn{1}{c|}{52.59} & \multicolumn{1}{c|}{49.36} & 47.19 \\ 
  & Full finetuning+MARC & \multicolumn{1}{c|}{52.38} & \multicolumn{1}{c|}{66.86} & \multicolumn{1}{c|}{55.36} & 44.83 & \multicolumn{1}{c|}{54.99} & \multicolumn{1}{c|}{52.38} & 50.03 \\
  & Full finetuning+CRT & \multicolumn{1}{c|}{\textbf{56.36}} & \multicolumn{1}{c|}{66.71} & \multicolumn{1}{c|}{\textbf{57.96}} & \textbf{51.63} & \multicolumn{1}{c|}{\textbf{60.21}} & \multicolumn{1}{c|}{\textbf{56.36}} & \textbf{54.73} \\
  \midrule

  \multirow{10}{*}{Places-LT}  & Zero-shot & \multicolumn{1}{c|}{32.17} & \multicolumn{1}{c|}{36.23} & \multicolumn{1}{c|}{30.43} & \textbf{37.89} & \multicolumn{1}{c|}{34.38} & \multicolumn{1}{c|}{32.17} & 30.85 \\
  & Linear probing & \multicolumn{1}{c|}{8.52} & \multicolumn{1}{c|}{23.50} & \multicolumn{1}{c|}{0.20} & 0.00 & \multicolumn{1}{c|}{7.03} & \multicolumn{1}{c|}{8.52} & 4.81 \\
  & Full finetuning & \multicolumn{1}{c|}{32.08} & \multicolumn{1}{c|}{47.32} & \multicolumn{1}{c|}{28.66} & 11.80 & \multicolumn{1}{c|}{41.55} & \multicolumn{1}{c|}{32.08} & 30.40 \\
  & Decoder+SoftMax & \multicolumn{1}{c|}{26.77} & \multicolumn{1}{c|}{43.73} & \multicolumn{1}{c|}{20.72} & 9.38 & \multicolumn{1}{c|}{39.68} & \multicolumn{1}{c|}{26.77} & 24.84 \\
  & Decoder+Balanced SoftMax & \multicolumn{1}{c|}{  35.19} & \multicolumn{1}{c|}{38.31} & \multicolumn{1}{c|}{34.84} & 30.24 & \multicolumn{1}{c|}{37.25} & \multicolumn{1}{c|}{35.19} & 34.20 \\ 
  & Decoder+MARC & \multicolumn{1}{c|}{35.33} & \multicolumn{1}{c|}{34.69} & \multicolumn{1}{c|}{36.24} & 34.42 & \multicolumn{1}{c|}{36.83} & \multicolumn{1}{c|}{35.33} & 34.53 \\
  & Decoder+CRT & \multicolumn{1}{c|}{34.23} & \multicolumn{1}{c|}{41.86} & \multicolumn{1}{c|}{36.17} & 15.69 & \multicolumn{1}{c|}{39.58} & \multicolumn{1}{c|}{34.23} & 32.63 \\
  & Full finetuning+Balanced SoftMax & \multicolumn{1}{c|}{\textbf{37.81}} & \multicolumn{1}{c|}{42.42} & \multicolumn{1}{c|}{\textbf{38.41}} & 27.93 & \multicolumn{1}{c|}{39.75} & \multicolumn{1}{c|}{\textbf{37.81}} & \textbf{37.21} \\ 
  & Full finetuning+MARC & \multicolumn{1}{c|}{37.13} & \multicolumn{1}{c|}{46.57} & \multicolumn{1}{c|}{38.09} & 17.51 & \multicolumn{1}{c|}{41.17} & \multicolumn{1}{c|}{37.13} & 35.95 \\
  & Full finetuning+CRT & \multicolumn{1}{c|}{33.51} & \multicolumn{1}{c|}{\textbf{47.81}} & \multicolumn{1}{c|}{30.77} & 13.39 & \multicolumn{1}{c|}{\textbf{45.35}} & \multicolumn{1}{c|}{33.51} & 33.04 \\
 \bottomrule
\end{tabular}
}
\end{table*}

\subsection{Main Results}
\tablename~\ref{tab-imagenet}, \ref{tab-inat} and \ref{tab-places} present the results on ImageNet-LT, iNaturalist18, and Places-LT datasets, respectively.
It should be noted that we were unable to provide the results of CoOp on the iNaturalist18 dataset due to the large number of classes (8142), which rendered it computationally infeasible to run CoOp, even with NVIDIA A100 80G hardware resources.
Other than ViT of CLIP, in \tablename~\ref{tab-laion}, we further provide some results using ViT of Laion-CLIP to show the generalizability of our decoder approach combining VLMs and imbalanced learning methods. Our findings are:

\textbf{1) VLMs exhibit strong zero-shot prediction abilities on ImageNet-LT, but domain-specific algorithms are necessary for achieving high performance on more diverse and fine-grained datasets such as iNaturalist18.} Specifically, on iNaturalist18, the CLIP and Laion-CLIP methods exhibit overall accuracies of 5.45\% and 3.82\%, respectively, which are significantly lower than those achieved by VLMs in conjunction with supervised methods. The reason can be that iNaturalist18 consists of a broad array of classes(more than 8000), each containing images of numerous species of plants and animals, some of which can be extremely similar visually. The similarity pattern presents a unique challenge for models like CLIP. The images on the Places-LT and ImageNet-LT datasets do not have such a high degree of inter-class visual similarity. By incorporating a lightweight decoder and imbalanced algorithms, CLIP achieved significant improvements in average overall accuracy compared to the zero-shot accuracy on three diverse datasets, namely ImageNet-Lt, iNaturalist18, and Places-LT, with performance gains ranging from 6.17\% to 69.82\%.

\textbf{2) The lightweight decoder is necessary for extracting relevant features and producing compact representations for tail classes, and it is essential for achieving high performance in the imbalanced setting.}
The results demonstrate that linear probing, which uses only the pre-trained encoder and a linear classifier, produces results that are only slightly better than zero-shot learning. The improvements achieved on iNaturalist18 and Places-LT are only 5.08\% and 0.49\%, respectively, which are significantly lower than the 69.82\% and 6.17\% improvements brought by the decoder. It is worth noting that on ImageNet-LT, linear probing even performs 3.16\% worse than zero-shot learning. Note that the full finetuning even perform similarly with linear probing on ImageNet-LT and Places-LT. Though full finetuning is much better than linear probing on iNaturalist, the performance of full finetuning is still much worse that that of imbalanced learning with a lightweight decoder.

\textbf{3) The decoder with SoftMax outperforms prompt tuning.}
Specifically, our decoder method with SoftMax achieves an overall accuracy of 75.05\% and an F1 score of 73.96\% on ImageNet-LT, outperforming CoOp by 14.75\% and 14.96\%, respectively. On Places-LT, our method achieves an overall accuracy of 40.15\% and an F1 score of 38.60\%, outperforming CoOp by 14.05\% and 14.60\%, respectively. One possible reason is that the training objective of our decoder with supervised learning method is more focused and aligned with the ultimate goal of the classification task. Moreover, CoOp requires heavy computational resources, which can make it infeasible to use when the number of classes is large, as is the case with the iNaturalist18 dataset.
This may limit the applicability of prompt tuning approaches for large number of classes.

\subsection{Analysis of Backbones}
\label{sec-exp-backbone}

Will larger backbones improve the performance?
In Table~\ref{tab-backbones}, we explore two variations of the CLIP backbone and report overall accuracy on all datasets with different imbalanced methods.
The ViT-L14 model has 443M parameters and ViT-B16 variant has 85M parameters.
As the number of parameters scale up, the overall accuracies on all datasets get improved significantly with both one-stage and two-stage imbalanced methods. 

Besides, we conduct experiments on CLIP-ResNet101. The training hyperparameters are the same with the main experiments except that we use a batch size of 128 for all ResNet experiments. As illustrated in \figurename~\ref{tab-resnet}, the performances of imbalanced methods are much better than Zero-shot, Linear-probing and full finetuning. Note the Transformer-like architecture of the decoder might not be feasible when applied to a ResNet~\cite{he2016deep} and their performances are worse than full finetuning with imbalanced methods. The reason might be that the blocks in our decoder closely resemble those in ViT, making their integration more straightforward. Furthermore, the features extracted by ResNet may not be suitable for a Transformer-based Decoder in imbalanced settings. Future research may focus on developing or modifying methods to suit different architectures and tasks for improved performance.

\begin{table*}[t!]
  \centering
  \caption{GPU memory usage during training on ImageNet-LT dataset, all tested with batch size set to 32.} 
  \label{tab-gpumem}
\resizebox{.5\textwidth}{!}{
\begin{tabular}{l|c|c}
\toprule
\multicolumn{1}{c|}{Method}                                    & Backbone & GPU Memory (MiB) \\ \midrule
\multicolumn{1}{c|}{\multirow{2}{*}{CLIP with Linear Probing}} & ViT-B16  & 3,796    \\
\multicolumn{1}{c|}{}                                          & ViT-L14  & 8,206    \\ \midrule
\multirow{2}{*}{CLIP with Decoder}                             & ViT-B16  & 4,456    \\
                                                               & ViT-L14  & 9,330    \\ \midrule
\multirow{2}{*}{CoOp(M=16, 1-shot, end)}                       & ViT-B16  & 20,974   \\
                                                               & ViT-L14  & 30,557   \\ \bottomrule
\end{tabular}
}
\end{table*}%

\begin{table*}[t!]
  \centering
  \caption{Training cost and carbon footprint of different methods and backbones on a single dedicated GPU server. We conducted tests on ImageNet-LT and measured average power consumption with {\ttfamily ipmitool} to calculate carbon footprint.} 
  \label{tab-cost}
\resizebox{.9\textwidth}{!}{
\begin{tabular}{c|c|c|c|c|c|c}
\toprule
\multicolumn{1}{c|}{Method} &
  Backbone &
  \makecell{Training Speed \\ (sec./iter.)} &
  \makecell{GPU Hour \\ ($h$)} &
  \makecell{Average Power \\ ($W$)} & \makecell{Total Power \\ ($KWh$)} & \makecell{Carbon Emitted \\ ($kgCO_{2}eq$)} \\ \midrule
\multirow{2}{*}{\makecell{One-Stage \\ Algorithms}} & ViT-B16 & 0.30 & 0.68 & 744.00 & 0.51 & 0.28 \\
                                      & ViT-L14 & 0.99 & 2.25 & 793.25 & 1.78 & 0.97 \\ \midrule
\multirow{2}{*}{\makecell{Two-Stage \\ Algorithms}} & ViT-B16 & 0.27 & 0.61 & 734.15 & 0.45 & 0.24 \\
                                      & ViT-L14 & 0.89 & 2.05 & 789.70 & 1.62 & 0.88 \\ \midrule
\multirow{2}{*}{Linear Probing}       & ViT-B16 & 0.23 & 0.52 & 690.55 & 0.36 & 0.20 \\
                                      & ViT-L14 & 0.84 & 1.91 & 783.45 & 1.50 & 0.81 \\ \midrule
CoOp(1-shot, M=16)                                  & ViT-B16 & 0.36 & 0.43 & 680.95 & 0.29 & 0.16 \\ \bottomrule
\end{tabular}}
\end{table*}%

\subsection{Analysis of Pre-training Data}
Will more pre-trained data lead to better performance?
Table~\ref{tab-laion} presents a comparison of the performances of CLIP (400M pre-trained data) and Laion-CLIP (2B pre-training data)~\citep{schuhmannlaion} on iNaturalist18 and Places-LT datasets.
Both models were evaluated using zero-shot learning and different imbalanced learning methods including Balanced SoftMax, DisAlign, and MARC.
First, it is surprising to find that Laion-CLIP achieves similar performance on iNaturalist, where CLIP performs extremely poor.
Second, it is shown that tuned CLIP is generally better than Laion-CLIP when combined with imbalanced methods, indicating that downstream tuning may be better than just increasing pre-training data.
Specifically, on iNaturalist18, the overall accuracy of tuned CLIP is usually 10\% higher than that of tuned Laion-CLIP.
One possible reason is that as the dataset size increases, the model becomes increasingly dominated by the most common classes, and CLIP's pre-training on a smaller dataset may better capture the representation of the tail classes.

\subsection{Analysis on Training Cost}

We analyze training costs from hardware requirement and power consumption.

On a NVIDIA A100 80G GPU, we experimented with various methods for training on ImageNet-LT. Results from \tablename~\ref{tab-gpumem} demonstrate that imbalanced learning techniques that employ a decoder only lead to a 13\% to 17\% increase in VRAM usage during training compared to linear probing. We ensured a VRAM usage of less than 10GB, enabling us to train and evaluate these methods on any consumer GPU. However, running CoOp on ViT-L14 requires a datacenter GPU, and at least an NVIDIA RTX 3090 or 4090 is necessary to execute the ViT-B16 variant of CLIP. We were unable to conduct experiments with CoOp on the iNaturalist dataset due to an Out-Of-Memory issue with ViT-B16 on a 1-shot setting.

The cost of training can include significant components such as power consumption and carbon footprint. We tested on a dedicated server with an RTX 4090 GPU and measured average power consumption using {\ttfamily ipmitool} during the process. We estimated carbon emission using the average carbon intensity (0.544 $kgCO_{2}/KWh$) in China. \tablename~\ref{tab-cost} demonstrates that imbalanced algorithms are more power-efficient than linear probing, particularly two-stage algorithms. However, CoOp, benefiting from its few-shot setting, consumed the least power and only emitted 67\% of the carbon compared to imbalanced methods.



\section{Limitation}

This work has several limitations.
First, we only use a limited number of VLMs such as CLIP and Laion-CLIP. Other VLMs~\citep{yu2022coca,dehghani2023scaling} may perform differently on long-tailed data and yield different results.
Second, this study only evaluates the performance on three popular long-tailed datasets, but there are still many other datasets with different characteristics that have not been explored.
Third, we do not use the text encoder for VLMs in the corporation with imbalanced algorithms and future exploration of adopting text encoder should be explored as well.

\section{Conclusion}

Our study highlights the significance of imbalanced learning in face of vision-language large models.
We show that current VLMs are not perfect in imbalanced tasks, e.g., CLIP only achieves 5\% on iNaturalist dataset.
For improvement, we introduced the decoder modification and the incorporation of imbalanced learning algorithms.
For instance, by adopting imbalanced learning, CLIP can improve from 5\% to 69\% on iNaturalist dataset.
We further show that pre-training data is weakly correlated with the performance, while the backbone size influences the results.
Finally, the incorporation with imbalanced algorithms do not significantly introduce computation burden.

In the future, we plan to explore the use of VLMs and imbalanced methods on more challenging datasets. Besides, we will try to explore VLMs in other settings with fewer labels or in unsupervised scenarios. We hope that our work will inspire further research in this direction and contribute to the development of more powerful and effective computer vision.

\bibliography{sn-bibliography}

\newpage
\begin{appendices}

\section{Algorithm Flow}
\label{sec-append-algo}

The algorithm of incorporating imbalanced learning algorithms in shown in Algo.~\ref{alg:two stage training}.

\section{Dataset}
\label{sec-append-data}

\begin{table*}[t!]
\centering
\caption{The detailed statistics of datasets.} 
\label{tab-dataset}
\begin{tabular}{l|ccc}
\toprule
Dataet        & \#Class & Training size & Test size  \\ \midrule
ImageNet-LT   & 1,000    & 115,846                   & 50,000         \\
Places-LT     & 365     & 62,500                   & 7,300        \\
iNaturalist18 & 8,142    & 437,513                  & 24,426     \\ \bottomrule                   
\end{tabular}
\end{table*}
The detailed statistics of datasets are shown in \tablename~\ref{tab-dataset}.

\begin{algorithm*}[htbp]
\caption{The training procedure of incorporating LVMs for different imbalanced learning algorithms.}
\label{alg:two stage training}
\begin{algorithmic}[1]
\State{\textbf{Input:} A batch of the training dataset $\mathcal{D}_s=\{(\mathbf{x}_i,y_i)\}_{i=1}^s$, encoder $\tilde{\theta}_e$, decoder $\theta_d$, classifier $\theta_c$, number of classes $K$, class-specific loss weight $\textcolor{red}{w}$, class-specific logit bias $\textcolor{blue}{\delta}$ and additional trainable parameters to adjust the original logits \textcolor{brown}{$\theta_a$}}. Note that the representation function is composed of $\tilde{\theta}_e$ and $\theta_d$ and parameters with $\tilde{\cdot}$ are fixed during training. 
\State{\textbf{First-stage} method: \emph{using instance-balanced sampling and the cross entropy loss.}}
\While{not reach the maximum iteration}
\State{Compute the loss and update the model parameters.}
\If{Using Cross-Entropy Loss}
        \State $\ell(\mathcal{D}_s;\tilde{\theta}_e,\theta_d,\theta_c) = -\frac{1}{s} \sum_{i=1}^s \log \left(\frac{\exp (\eta_{y_i})}{\sum_{j=1}^K  \exp(\eta_j)}\right),$ where $\eta_j$ is classification score of class $j$. 
\Else{ Using Class-Balanced Loss}
       \State $\ell(\mathcal{D}_s;\tilde{\theta}_e,\theta_d,\theta_c) = -\frac{1}{s} \sum_{i=1}^s \textcolor{red}{w_{y_i}} \log \left(\frac{\exp (\eta_{y_i}+\textcolor{blue}{\delta_{y_i}})}{\sum_{j=1}^K  \exp(\eta_j+\textcolor{blue}{\delta_{j}})}\right),$ where $\eta_j$ is classification score of class $j$. 
\EndIf

\EndWhile
\State{\textbf{Second stage}: calibrate the model trained in the first stage.}
\While{not reach the maximum iteration}
\State{\emph{Use instance-balanced sampling~\citep{zhang2021distribution,wang2022margin} or class-balanced sampling~\citep{kang2019decoupling} methods.}}
\State{Compute the loss and update the model parameters.}
\If{Using Cross-Entropy Loss}
        \State $\ell(\mathcal{D}_s;\tilde{\theta}_e,\tilde{\theta}_d,\theta_c,\textcolor{brown}{\theta_a}) = -\frac{1}{s} \sum_{i=1}^s \log \left(\frac{\exp (\eta_{y_i})}{\sum_{j=1}^K  \exp(\eta_j)}\right),$ where $\eta_j$ is classification score of class $j$. 
\Else{Using Class-Balanced Loss}
       \State $\ell(\mathcal{D}_s;\tilde{\theta}_e,\tilde{\theta}_d,\theta_c,\textcolor{brown}{\theta_a}) = -\frac{1}{s} \sum_{i=1}^s \textcolor{red}{w_{y_i}} \log \left(\frac{\exp (\eta_{y_i}+\textcolor{blue}{\delta_{y_i}})}{\sum_{j=1}^K  \exp(\eta_j+\textcolor{blue}{\delta_{j}})}\right),$ where $\eta_j$ is the logit of class $j$. 
\EndIf
\EndWhile
\State{\textbf{Return:} Model parameters $\theta_r, \theta_c$, $\textcolor{brown}{\theta_a}$.}
\end{algorithmic}
\end{algorithm*}

\section{Imbalanced Algorithms}
\label{intro-algs}

In this section, we give a brief introduction to the used imbalanced methods.



\textbf{Class Balanced Re-Weighting} Class Balanced Re-Weighting (CBW) assigns loss weights to each instance in the dataset based on the class distribution, such that each class has an equal contribution to the overall loss function during training, which allows the model to give more importance to the minority class.

\textbf{LDAM Loss} Label-Distribution-Aware Margin (LDAM) Loss~\citep{LDAM} aims to improve the performance of the model on imbalanced datasets by considering the distribution of labels in the data. This loss function adds a margin term to the traditional cross-entropy loss, which prevents the model from being biased towards the majority class. The margin term  calculated based on the class distribution in the dataset.

\textbf{Focal Loss} Focal Loss~\citep{focal} assigns higher weights to hard-to-classify samples which has low confidence in prediction, making them more important in the training process, while reducing the contribution of easy-to-classify samples with high confidence. 

\textbf{Balanced Softmax Loss} Balanced Softmax Loss~\citep{ren2020balanced} propose an unbiased extension of Softmax called Balanced Softmax, which accommodates the label distribution shift between training and testing. It can minimize the generalization bound in the imbalanced settings.

\textbf{LADE Loss}  LAbel distribution DisEntangling (LADE) Loss~\citep{hong2021disentangling} formulates imbalanced classification as a label shift problem where the target and source label distributions are different, and identifies the entanglement between the source label distribution and the model prediction as a significant hurdle. LADE loss is based on the optimal bound of Donsker-Varadhan representation to directly disentangle the source label distribution from the model prediction in the training phase. 


\textbf{CRT and LWS} ~\citep{kang2019decoupling} focuses on exploring the impact of representation strategies and classifier strategies and finds that data imbalance may not be a major issue in learning high-quality representations. They demonstrate that it is possible to achieve strong imbalanced classification ability by adjusting only the classifier, even when the representations are learned using the simplest instance-balanced (natural) sampling. \citep{kang2019decoupling} proposes a straightforward approach called Classifier Re-Training (CRT) which re-trains the re-initialized classifier with class-balanced sampling and fixed representations. Besides Learnable weight scaling (LWS) can also improve the performance of imbalanced classification by re-scaling of the magnitude for the weight matrices for each class in the classifier.

\textbf{Disalign} Disalign~\citep{zhang2021distribution} is also a two stage algorithms like CRT and LWS. It keeps both the representation and the classifier fixed and develops an adaptive calibration function to adjust the classification scores by adding class specific extra classifier and instance specific confidence layer. 

\textbf{MARC} In~\citep{wang2022margin}, the relationship between the margins and logits is examined, and a positive correlation is observed between the biased margins and biased logits. To address this issue, MARgin Calibration function (MARC) with only $2K$ trainable parameters ($k$ is the number of classes) is proposed to dynamically calibrates the biased margins to obtain unbiased logits with both the representation and the classifier fixed.

\end{appendices}

\end{document}